\documentclass[11pt]{article}

\usepackage[truedimen,margin=25truemm]{geometry}

\usepackage{graphicx} 
\usepackage{epsfig} 
\usepackage{subfigure} 

\usepackage{amsmath}
\usepackage{amsthm}
\usepackage{amssymb}

\usepackage{algorithm}
\usepackage{algorithmic}

\usepackage{color}

\usepackage{ulem}

\usepackage{booktabs}

\usepackage{hyperref}

\usepackage{graphicx} 
\usepackage{epsfig}
\usepackage{subfigure}

\usepackage{amssymb}
\usepackage{amsmath}
\usepackage{amsthm}
\usepackage{mathtools}
\usepackage[raggedright]{titlesec}

\def\vec#1{\mbox{\boldmath $#1$}}
\def\mat#1{\mbox{\bf #1}}

\title{Consistency-aware and Inconsistency-aware Graph-based Multi-view Clustering}

\author{Mitsuhiko Horie \thanks{Department of Communications and Computer Engineering, School of Fundamental Science and Engineering, WASEDA University, 3-4-1 Okubo, Shinjuku-ku, Tokyo 169-8555, Japan (e-mail: kidsgoldppp@akane.waseda.jp) } \and Hiroyuki Kasai \thanks{Department of Communications and Computer Engineering, School of Fundamental Science and Engineering, WASEDA University, 3-4-1 Okubo, Shinjuku-ku, Tokyo 169-8555, Japan (e-mail: hiroyuki.kasai@waseda.jp)}}

\begin{document}

\maketitle

\begin{abstract}
Multi-view data analysis has gained increasing popularity because multi-view data are frequently encountered in machine learning applications. A simple but promising approach for clustering of multi-view data is multi-view clustering (MVC), which has been developed extensively to classify given subjects into some clustered groups by learning latent common features that are shared across multi-view data. Among existing approaches, graph-based multi-view clustering (GMVC) achieves state-of-the-art performance by leveraging a shared graph matrix called the unified matrix. However, existing methods including GMVC do not explicitly address inconsistent parts of input graph matrices. Consequently, they are adversely affected by unacceptable clustering performance. To this end, this paper proposes a new GMVC method that incorporates consistent and inconsistent parts lying across multiple views. This proposal is designated as CI-GMVC. Numerical evaluations of real-world datasets demonstrate the effectiveness of the proposed CI-GMVC \footnote{This paper has been accepted in EUSIPCO2020 \cite{horie_eusipo_2020}.}.
\end{abstract}

\section{Introduction}
\label{Sec:Intro}
Many machine learning applications such as image classification, social networks, chemistry, signal processing, web analysis, item recommendation, and bioinformatics analysis usually exhibit some {\it structured data}, which might include trees, groups, clusters, paths, sequences \cite{Narimatsu2015,narimatsu2017state,Narimatsu_ICASSP_2020}, and graphs \cite{Hashimoto_ICASSP_2020}. Recent advances in information-retrieval technologies enable collection of such structured data with {\it heterogeneous} features from {\it multi-view} data. For example, each web page includes two views of text and images. Image data include multiple features such as color histograms and frequency features of wavelet coefficients. The emergence of such multi-view data has raised a new question: how can such multiple sets of features for individual subjects be integrated into data analysis tasks? This question motivates a new paradigm, called {\it multi-view learning}, for data analysis with multi-view feature information. Multi-view learning fundamentally makes use of common or consensus information that is presumed to exist across multi-view data to improve data analysis task performance. One successful subcategory of multi-view learning is {\it multi-view clustering} (MVC), which classifies given subjects into subgroups based on similarities among subjects \cite{Chao_arXiv_2017}. Although various approaches have been proposed in this category, graph-based multi-view clustering (GMVC) has recently garnered increasing attention: it has demonstrated state-of-the-art performance for numerous applications \cite{Saha_PReMI_2013,Nie_IJCAI_2016,How_IEEEKDE_2017_s,Nie_IJCAI_2017,Tao_ACML_2017,Zhuge_IEEETKDE_2017,Nie_AAAI_2017,Nie_IEEETIP_2018,Zhan_IEEETC_2018}. Fundamentally, GMVC originates from single-view spectral clustering methods. It performs clustering tasks by exploiting consensus features across input {\it graph matrices} such as adjacency matrices. Among them, some works outperform other methods by seeking a consensus matrix called a {\it unified matrix} from the input graph matrices with adaptive weights such that the unified matrix directly represents the final clustering result \cite{Tao_ACML_2017,Nie_IJCAI_2016,Nie_IJCAI_2017,Nie_IEEETIP_2018, Wang_KBS_2019_s}.

Most existing methods achieve good performance by exploiting the {\it consensus part} across multiple views. However, each view might include an {\it inconsistent part} that does not appear in other views because of noise and outliers. Therefore, this inclusion might lead to severe degradation of downstream clustering performance. Nevertheless, they do not deal explicitly with such inconsistent parts of the input graph matrices. Consequently, they offer only unacceptable performance.

To alleviate this issue, this paper presents a proposal of graph-based multi-view clustering, which particularly incorporates a consistent part and an inconsistent part lying across multiple views. More concretely, we separate the input graph matrices collected from multi-view data into a consistent part and an inconsistent part by orthogonality constraints. This conceptualization shares similar ideas with some recent works \cite{Bojchevski_KDD_2017, Liang_ICDM_2019_s}. Derivation of the unified matrix is therefore more efficient and more robust than existing GMVC methods, leading to improvement of the clustering performance. We designate this proposed algorithm as CI-GMVC. Numerical evaluations conducted with comparison to state-of-the-art multi-view clustering methods reveal the effectiveness of the proposed CI-GMVC on several real-world datasets. The source code is available at \url{https://github.com/hiroyuki-kasai}.

\section{Preliminary explanation}

This subsection first presents a summary of the notation used in the remainder of this paper. Subsequently, we briefly introduce multi-view clustering and specifically address graph-based multi-view clustering, which are basic techniques of the proposed algorithm: CI-GMVC.

\subsection{Notation}

We represent scalars as lower-case letters $(a, b, \ldots)$, vectors as bold lower-case letters $(\vec{a}, \vec{b}, \ldots)$, and matrices as bold-face capitals $(\mat{A}, \mat{B}, \ldots)$. The $i$-th element of \vec{a} and the element at the $(i,j)$ position of \mat{A} are represented respectively as $a_i$ and $A_{ij}$. Also, the vector of the $i$-th row vector $A_{i:}$ and the vector of the $j$-th column vector $A_{:j}$ are denoted respectively as $\vec{a}_i$ and $\vec{a}^j$. $\vec{1}_d$ is used for the $d$-dimensional vector of ones, and $\mat{I}_d$ is the identity matrix of size $d \times d$. $\mathbb{R}_{+}^{n \times m}$ represents a nonnegative matrix of size ${n \times m}$. Additionally, $\mat{A}\geq 0$ and $\mat{A}> 0$ respectively represent $A_{ij}\geq0$ and $A_{ij}>0$ for all $(i,j)$. Furthermore, $\mat{A}\geq \mat{B}$ and $\mat{A}> \mat{B}$ represent $A_{ij}\geq B_{ij}$ and $A_{ij}> B_{ij}$ for all $(i,j)$. Operators ${\rm Tr}(\cdot)$ and $(\cdot)^T$ respectively stand for the matrix trace and transpose. Operator $\mathrm{max}(\mat{A},\mat{B})$ outputs $A_{ij}$ when $A_{ij}\geq B_{ij}$ at the $(i,j)$-th position, and $B_{ij}$ otherwise. Operator $\mathrm{vec}(\mat{A})$ performs the vectorization of \mat{A}. Operator $\mathrm{diag}(\vec{a})$ represents \mat{A} of which diagonal and off-diagonal elements are \vec{a} and zeros, respectively. Regarding multi-view data, $N$, $V$, and $C$ respectively represent the number of sample data, views, and classes. The dimensions of sample data of the $c\ (\in [C])$-th class in the $v\ (\in [V])$-th view are denoted as $d_v$.

\subsection{Multi-view clustering: MVC}

Multi-view learning fundamentally makes use of common or consensus information across multi-view data \cite{Xu_arXiv_2013,Sun_NCA_2013, Zhao_IF_2017,Li_IEEETranKD_2019}. {\it Multi-view discriminant learning}, a supervised learning technique, has been studied extensively \cite{Rosipal_SLSFS_2005,Hotelling_1936_Biometrika,Thompson_2005_CCA,Tenenbaum_NC_2000,Sharma_CVPR_2012,Cao_2016_IT_s,Xu_2019_IEEETranCyber,Kasai_ICASSP_2020}. It generally originates from single-view linear discriminant analysis such as Fisher linear discriminant analysis (LDA or FDA) \cite{Hastie_2009_StatLearnBook}. Regarding unsupervised learning techniques, however, {\it multi-view clustering} (MVC) clusters given subjects into several groups such that the points in the same group are similar and the points in different groups are dissimilar to one another by combining multi-view data \cite{Chao_arXiv_2017}. One naive approach of MVC is to perform a single-view clustering method against {\it concatenated} features collected from different views. However, this approach might fail when higher emphases are put to certain specific views than to others. Consequently, this category of research has attracted more attention, to include multi-view subspace clustering that learns common coefficient matrices \cite{Wang_IJCAI_2016, Yin_CIKM_2015}, multi-view nonnegative matrix factorization clustering that learns common indicator matrices \cite{Akata_CVWW_2011}, multi-view $k$-means \cite{Cai_IJCAI_2013}, multi-kernel based multi-view clustering \cite{Guo_ICPR_2014} and CCA based multi-view clustering \cite{Blaschko_CVPR_2008}.

\subsection{Graph-based multi-view clustering: GMVC}
\label{Sec:GMVC}
Different from the approaches described above, graph-based MVC (GMVC) learns common eigenvector matrices or shared matrices, and empirically demonstrates state-of-the-art results in various applications. General steps consist of (i) generating an input graph matrix, called SIG, (ii) generating the graph Laplacian matrix, (iii) computing the embedding matrix, and (iv) performing clustering into groups using an external clustering algorithm. These steps are shared with the normalized cut  \cite{Shi_IEEEPAMI_2000} and the spectral clustering \cite{Mohar_GTCA_1991}. Furthermore, GMVC is related closely to multi-view spectral clustering \cite{Xia_IEEESMCB_2010, Kumar_NIPS_2011,Xia_AAAI_2014,Lu_IEEETIP_2016,Feng_SC_2017,Wang_IEEETNNLS_2018,Zong_AAAI_2018}. Recently, some works of GMVC address the unified matrix $\mat{U}(\in \mathbb{R}_+^{N \times N})$ with different effects of multiple graph matrices as \cite{Tao_ACML_2017,Nie_IJCAI_2016,Nie_IJCAI_2017,Nie_IEEETIP_2018, Wang_KBS_2019_s}. More noteworthy is that, whereas many MVC methods rely on an external clustering algorithm after learning, they seek a unified graph matrix such that it {\it internally} indicates cluster information. 

More specifically, considering an adaptive weight $\vec{\alpha} = [\alpha_1, \ldots, \alpha_v, \ldots, \alpha_V]^T (\in \mathbb{R}^{V})$ on all $V$ SIG matrices $\{\mat{S}_1, \ldots, \mat{S}_v, \ldots, \mat{S}_V\} (\in \mathbb{R}^{N \times N}_+)$, the following minimization problem is formulated with respect to \mat{U}:

\begin{eqnarray}
\label{Eq:obj_fun_1}
\min_{ \mat{\scriptsize U}} && \sum_{v=1}^{V}\alpha_v\|\mat{U}-\mat{S}_v\|_F^2\notag\\
{\rm subject\ to}&& U_{ij} \geq 0, \vec{1}^T \vec{u}_i= 1,
\end{eqnarray}
where $\vec{\alpha}$ represents the weight vector $\vec{\alpha}$, calculated as $\alpha_v=1/(2\sqrt{\|\mat{U}-\mat{S}_v\|_F^2})$, as in \cite{Wang_KBS_2019_s,Nie_IJCAI_2017,Nie_IEEETIP_2018}.
Furthermore, the graph Laplacian matrix $\mat{L}_{\mat{\scriptsize U}} (\in \mathbb{R}^{N \times N})$ of \mat{U} is introduced such that \mat{U} directly produces the clustering result without relying on external clustering methods. For this purpose, acknowledging that \mat{U} can be partitioned into $C$ groups directly when ${\rm rank(\mat{L}_{\mat{\scriptsize U}})}$ is equal to $n-C$ \cite{Dhillon_SIGMOD_2001}, and using Fan's theorem \cite{Fan_PNAS_1949}, the following formulation is proposed: 

\begin{eqnarray}
\label{obj_fun_2}
	\min_{\mat{\scriptsize U},\mat{\scriptsize F},\vec{\scriptsize \alpha}}&& \sum_{v=1}^{V}\alpha_v 
	\|\mat{U}-\mat{S}_v \|_F^2 + 2\lambda \cdot {\rm Tr}(\mat{F}^T\mat{L}_{\mat{\scriptsize U}}\mat{F})\notag\\
	{\rm subject\ to}&& U_{ij}\geq0, \vec{1}^{T}\vec{u}_i=1,\mat{F}^{T}\mat{F}=\mat{I}_C,
\end{eqnarray}
where $\lambda (>0)$ is a regularization parameter. $\mat{F} (\in \mathbb{R}^{N\times C})$ is an embedding matrix, which lies on the orthogonal matrix, i.e., the Stiefel manifold ${\rm St}(p,d)$; the Riemannian submanifold of orthonormal matrices $\mathcal{M}\!=\!\{\mat{X}\! \in\! \mathbb{R}^{d \times p}\!:\! \mat{X}\mat{X}^T\!=\!\mat{I}_p\}$.

\section{Proposed CI-GMVC: Consistency-aware and inconsistency-aware GMVC}
As explained in {Section \ref{Sec:GMVC}}, GMVC considers consensus features across input graph matrices, i.e., SIG matrices, across multiple views. The hypothesis in this scheme relies on the assumption that all SIG matrices $\{\mat{S}_1, \ldots, \mat{S}_v, \ldots, \mat{S}_V\}$ share common features across multi-views with appropriate weights. Subsequently, \mat{U} can be obtained efficiently from these matrices. However, as explained in {Section \ref{Sec:Intro}}, the SIG matrices $\mat{S}_v$ are not perfect: they might be corrupted because of noisy input data, severe outliers, and latent fundamental inconsistent structures \cite{Bojchevski_KDD_2017}. Consequently, any analysis of these matrices might result in unacceptable clustering results. Therefore, it is necessary to handle inconsistent parts among multi-view data to avoid lower quality of subsequent clustering.

To alleviate this issue, we separate $\mat{S}_v$ into a consistent part and an inconsistent part that exist as mixed inside $\mat{S}_v$, and calculate \mat{U} {\it only} from the consistent parts of $\mat{S}_v$. More concretely, we assume that $\mat{S}_v$ consists of a consistent part $\mat{A}_v (\in \mathbb{R}^{N \times N})$ and an {\it nonnegative} inconsistent part $\mat{E}_v (\in \mathbb{R}_+^{N \times N})$ as $\mat{S}_v=\mat{A}_v+\mat{E}_v$ which follows\cite{Bojchevski_KDD_2017,Liang_ICDM_2019_s}.
Assuming further that the elements of $\mat{E}_v$ are {\it not shared} in other $\mat{E}_w (v \neq w)$, in other words, expecting that they are {\it element-wise orthogonal}, we assign a penalty as
\begin{eqnarray*}
	\sum_{v,w=1}^{V}\gamma\cdot {\rm vec}(\mat{E}_v)^T {\rm vec}(\mat{E}_w) = \sum_{v,w=1}^{V}\gamma\cdot{\rm Tr}(\mat{E}_v\cdot(\mat{E}_w)^{T}),
\end{eqnarray*}
where $\gamma (>0)$ is a weighting hyperparameter. Furthermore, a penalty of larger elements of $\mat{E}_v$ is regarded as stabilizing the optimization of the objective function as
\begin{equation*}
	 \sum_{v=1}^{V} \beta\| {\rm vec}(\mat{E}_v) \|^2_2 = \sum_{v=1}^{V}\beta\cdot{\rm Tr}(\mat{E}_v\cdot(\mat{E}_v)^{T}),
\end{equation*}
where $\beta (>0)$ is a weighting hyperparameter. Integrating the two penalties above yields the following penalty term.
\begin{equation*}
	\sum_{v,w=1}^{V}b_{vw}\cdot {\rm Tr}(({\bf S}_v-\mat{A}_v)\cdot(\mat{S}_w-\mat{A}_w)^T).
\end{equation*}

Therein, $b_{vw}$ represents the $(v,w)$-elements of $\mat{ B} (\in \mathbb{R}^{V\times V})$, of which diagonal and off-diagonal elements respectively correspond to $\beta$ and $\gamma$. 
Finally, we derive the objective formulation mathematically as presented below:
\begin{eqnarray}
\label{Eq:obj_fun}
	\min_{\scriptsize \substack{\mat{U},\mat{F},\vec{\alpha}\\\mat{A}_1,\ldots,\mat{A}_V}}&&\sum_{v=1}^{V}\alpha_v\|\mat{U}-\mat{A}_v\|_F^2+2\lambda{\rm Tr}(\mat{F}^T\mat{L}_{\mat{\scriptsize U}}\mat{F})+\sum_{v,w=1}^{V}b_{vw}\cdot {\rm Tr}(({\bf S}_v-\mat{A}_v)\cdot(\mat{S}_w-\mat{A}_w)^{T})\\
	{\rm subject\ to}&& U_{ij}\geq0, \vec{1}^{T}\vec{u}_i=1,\mat{F}^{T}\mat{F}=\mat{I}_C, {\bf S}_v\geq\mat{A}_v\geq0.\notag
\end{eqnarray}

As shown there, the differences one can note in relation to existing works are that the unified matrix \mat{U} is evaluated with $\mat{A}_v$, i.e., the consistent part of $\mat{S}_v$, and the inconsistent part $\mat{S}_v-\mat{A}_v$ are evaluated with $\mat{S}_w-\mat{A}_w (v\neq w)$ in terms of their mutual orthogonality. 

\section{Optimization Algorithm}
The objective function in (\ref{Eq:obj_fun}) is not jointly convex on all variables. Therefore, the alternating minimization algorithm is exploited to obtain the solutions. Note that, as mentioned earlier, $\vec{\alpha}$ is calculated as $\alpha_v=1/(2\sqrt{\|\mat{U}-\mat{A}_v\|_F^2})$, as in \cite{Wang_KBS_2019_s,Nie_IJCAI_2017,Nie_IEEETIP_2018}.

\subsection{Update of \mat{F} and \mat{U}}
The updates of \mat{F} and \mat{U} are similar to those of \cite{Wang_KBS_2019_s}. However, for the self-contained explanation, we briefly describe their update rules. The optimization problem in (\ref{Eq:obj_fun}) with respect to the unified matrix \mat{U} yields the following: 
\begin{eqnarray*}
\label{Eq:obj_fun_U}
	\min_{\mat{\scriptsize U}}&&\sum_{v=1}^{V}\alpha_v\|\mat{U}-\mat{A}_v\|_F^2
		+2\lambda\cdot{\rm Tr}(\mat{F}^T\mat{L}_{\mat{\scriptsize U}}\mat{F})\\
	{\rm subject\ to}&& U_{ij}\geq0,\vec{1}^{T}\vec{u}_i=1\notag.
\end{eqnarray*}

This is equivalent to the following minimization problem in terms of $\vec{u}_i$ for $i \in [N]$ as 
\begin{eqnarray}
\label{Eq:SolutionU}
	\min_{\vec{u}_i}&&\sum_{v=1}^{V}
	\| \vec{u}_i - (\vec{a}_v)_i + \frac{\lambda}{2V\alpha_v} \vec{p}_i \|_2^2\\
	{\rm subject\ to}&& U_{ij}\geq0,\vec{1}^{T}\vec{u}_i=1,\notag
\end{eqnarray}
where $\vec{p}_i=[(p_i)_1, (p_i)_2,\ldots,(p_i)_j,\ldots, (p_i)_N]^T (\in \mathbb{R}^{N})$ and where
$(p_i)_j$ is equal to $\| \vec{f}^i - \vec{f}^j\|_2^2$; $ \vec{f}^i (\in \mathbb{R}^{1 \times C})$ is the $i$-th row vector of \mat{F}. This problem is solvable as in {Section 5.3} in \cite{Wang_KBS_2019_s}. Finally, the optimization problem in (\ref{Eq:obj_fun}) in terms of \mat{F} under fixed \mat{U}, \mat{A}, \vec{\alpha} is 
\begin{equation}
\label{Eq:SolutionF}
\begin{split}
	\min_{\mat{\scriptsize F}} \quad{\rm Tr}(\mat{F}^T\mat{L}_{\mat{\scriptsize U}}\mat{F}), \quad {\rm subject\ to}\quad \mat{F}^{T}\mat{F}=\mat{I}_C.
\end{split}
\end{equation}

The solution is obtainable to calculate the $C$ eigenvectors of $\mat{L}_{\mat{\scriptsize U}}$, of which eigenvalues are the $C$ smallest ones \cite{Mohar_GTCA_1991}.

\subsection{Update of \mat{A}}
Keeping \mat{F}, \mat{U}, $\vec{\alpha}$ as fixed, the minimization problem about $\mat{A}_v$ is defined as
\begin{eqnarray*}
\label{Eq:obj_fun_A}
	\min_{\scriptsize \mat{A}_1,\ldots,\mat{A}_V}&&
	\sum_{v=1}^{V}\alpha_v\|\mat{U}-\mat{A}_v\|_F^2+\sum_{v,w=1}^{V}b_{vw}\cdot {\rm Tr}(({\bf S}_v\!-\!\mat{A}_v)\!\cdot\!(\mat{S}_w-\mat{A}_w)^{T})\\
	{\rm subject\ to}\ \ &&{\bf S}_v\geq\mat{A}_v\geq0.
\end{eqnarray*}

The first-order necessary optimality conditions of this problem are that its gradient with respect to $\mat{A}_v$ is expected to be zero, which means
\begin{eqnarray*}
	2\alpha_v(\mat{U}-\mat{A}_v)+\sum_{w=1}^{V}b_{vw}(-\mat{S}_w+\mat{A}_w) &=& \vec{0},
\end{eqnarray*}
for $v=1,2,\ldots,V$. Consequently, we obtain the following.
\begin{eqnarray}
\label{Eq:DeriveA}
	2\alpha_v\mat{A}_v+\sum_{w=1}^{V}b_{vw}\mat{A}_w&=&2\alpha_v\mat{U}+\sum_{w=1}^{V}b_{vw}\mat{S}_w.
\end{eqnarray}

Here, because the left-hand terms are represented as $b_{v1}\mat{A}_1 + \cdots +  (b_{vv}+2\alpha_v)\mat{A}_v + \cdots\ b_{vV}\mat{A}_V$, its vectorization form is represented as 
\begin{eqnarray*}
[b_{v1}\ b_{v2}\ \cdots\ (b_{vv}+2\alpha_v)\ \cdots\ b_{vV}]
\left(\begin{array}{c}{\rm vec}(\mat{A}_1)^T\\\vdots\\{\rm vec}(\mat{A}_V)^T\end{array}\right ).
\end{eqnarray*}

Adding all $V$ terms of $\mat{A}_v$ yields the following.

\begin{eqnarray*}
\left(
\begin{array}{ccccc}
 \!b_{11}\!+\!2\alpha_1\! & \!\!\cdots\!\! & \!\!\cdots\!\! & \!\!\cdots\!\! & b_{1V}\\
\!\!\vdots\!\! & \!\!\ddots\!\! & \!\!\cdots\!\!& \!\!\cdots\!\! & \!\!\vdots\!\!\\
b_{v1} & \!\!\cdots\!\! & b_{vv}\!+\!2\alpha_v & \!\!\cdots\!\! & b_{vV}\\
\!\!\vdots\!\! & \!\!\cdots\!\! & \!\!\cdots\!\!& \!\!\ddots\!\! & \!\!\vdots\!\!\\
b_{V1} & \!\!\cdots\!\! & \!\!\cdots\!\! & \!\!\cdots\!\! &  \!b_{VV}\!+\!2\alpha_V\!\! \!\\
\end{array}
\right )
\left(\begin{array}{c}\!\!{\rm vec}(\mat{A}_1)^T\!\!\\\vdots\\\!\!{\rm vec}(\mat{A}_V)^T\!\!\end{array}\right ).
\end{eqnarray*}

Denoting \textcolor{black}{$2 {\rm diag}(\vec{\alpha})+\mat{B}\ (\in \mathbb{R}^{V \times V})$} as $\mat{C}$, and the right-hand terms in (\ref{Eq:DeriveA}) as $\mat{H}_v\ (\in \mathbb{R}^{V \times V})$, we calculate $\mat{A}_v (v \in [V])$ as 

\begin{eqnarray}
\label{Eq:Update_A}
	\left(\begin{array}{c}{\rm vec}(\mat{A}_1)^T\\\vdots\\{\rm vec}(\mat{A}_V)^T\end{array}\right )&=&\mat{C}^+\cdot\left(\begin{array}{c}{\rm vec}(\mat{H}_1)^T\\\vdots\\{\rm vec}(\mat{H}_V)^T\end{array}\right ),
\end{eqnarray}
where $\mat{C}^+$ is the inverse or the pseudo-inverse matrix of \mat{C}.

Finally, considering the constraint of ${\bf S}_v\geq\mat{A}_v\geq0$, the final solution of $\mat{A}_v$ is obtainable by outputting $\mat{A}''_v$ as
\begin{eqnarray}
\label{Eq:Update_A_final}
	\mat{A}'_v=\max(\mat{A}_v,\mat{0}),\quad \mat{A}''_v=\min(\mat{A}'_v,{\bf S}_v).
\end{eqnarray}

The overall algorithm of the proposed CI-GMVC is summarized in {\bf Algorithm \ref{alg:proposed_alg}}.

\begin{algorithm}
\caption{CI-GMVC optimization algorithm}      
\label{alg:proposed_alg}
\begin{algorithmic}[1]       
\REQUIRE{Unified matrix \mat{U}.}
\ENSURE{SIG matrix ${\bf S}_1,\ldots,{\bf S}_V$, cluster number $C$, $\lambda$, $\beta$, $\gamma$.}
\STATE{Initialize $\alpha_v=1/V$ and $\mat{A}_v=\mat{S}_v$ for $v \in [V]$.}
\STATE{Initialize \mat{U} from \vec{\alpha} weighted summation of $\mat{S}_1, \dots, \mat{S}_V$.}
\STATE{Initialize \mat{F} using (\ref{Eq:SolutionF}).}
\STATE{Update $\vec{\alpha}$ as $\alpha_v=1/(2\sqrt{\|\mat{U}-\mat{A}_v\|_F^2})$.}
\STATE{Update \mat{U} with \mat{F}, \mat{A}, \vec{\alpha} fixed using (\ref{Eq:SolutionU}).}
\STATE{Update \mat{F} with \mat{U}, \mat{A}, \vec{\alpha} fixed using (\ref{Eq:SolutionF}).}
\STATE{Update \mat{A} with \mat{F}, \mat{U}, \vec{\alpha} fixed using  (\ref{Eq:Update_A}) and (\ref{Eq:Update_A_final}).}
\STATE{Repeat the steps presented above until \mat{U} converges or the predefined maximum number of iterations is reached.}
\end{algorithmic}
\end{algorithm}

\section{Numerical Evaluations}
This section presents empirical evaluation of the proposed CI-GMVC with some real-world datasets. We compare the proposed algorithm with state-of-the-art methods, which include 
Multi-view Spectral Clustering ({MSC})\footnote{Source code available at \url{https://github.com/frash1989/ELM-MVClustering/tree/master/RMSC-ELM}.}\cite{Xia_AAAI_2014}, 
Co-regularized Spectral Clustering ({CoregSC})\footnote{Source code available at \url{http://legacydirs.umiacs.umd.edu/~abhishek/code_coregspectral.zip}.} \cite{Kumar_NIPS_2011}, 
Multiple Graph Learning ({MGL})\footnote{Source code available at \url{http://www.escience.cn/people/fpnie}.} \cite{Nie_IJCAI_2016}, 
Multi-view Clustering with Graph Learning ({MCGL})\footnote{Source code available at \url{https://github.com/kunzhan/MVGL}.} \cite{Nie_IEEETIP_2018}, and
and Graph-based System ({GBS})\footnote{Source code available at \url{https://github.com/cswanghao/gbs}.}\cite{Wang_KBS_2019_s}. 
As for GBS and our proposed CI-GMVC methods, the SIG matrix is generated by following {Algorithm 1} in \cite{Wang_KBS_2019_s} with the number of neighbors $k=15$. The hyper-parameters for our proposed CI-GMVC are $\beta = 10^{-12}$ and $\gamma = 10^{-5}$, which are obtained from preliminary experiments.

Datasets summarized in {TABLE \ref{tbl:dataset}} are the following. 
The {BBC (BBC)} dataset\footnote{\url{http://mlg.ucd.ie/datasets/segment.html}.} includes news articles from the BBC news website. The number of articles is $685$. Each has one of five topical labels.
The {Newsgroup (NG)} dataset\footnote{\url{http://lig-membres.imag.fr/grimal/data.html}.} is collected from the $20$ News-group datasets, which has $500$ newsgroup documents. This has five topical labels.
The {WebKB} dataset\footnote{\url{https://linqs.soe.ucsc.edu/data}.} has four classes; it includes $203$ web-pages. Each web-page consists of the anchor text of the hyperlink, its title, and the page content.
The {One-hundred plant species leaves (100 leaf)} dataset\footnote{\url{https://archive.ics.uci.edu/ml/datasets/One-hundred+plant+species+leaves+data+set}.} includes three views of which one has $1600$ samples. Each belongs to one of the one hundred plant species.

\begin{table}[htbp]
\caption{Features of datasets used for this experiment.}

\label{tbl:dataset}
\begin{center}
\begin{tabular}{c|r|c|r|r|r|r|r} 
\hline
dataset &\multicolumn{1}{|c|}{$N$} & $V$ & \multicolumn{1}{|c}{$C$}  & \multicolumn{4}{|c}{dimensions of each view}\\
\cline{5-8}
&\multicolumn{1}{|c}{}&\multicolumn{1}{|c}{}&\multicolumn{1}{|c}{}& \multicolumn{1}{|c}{$d_1$} & \multicolumn{1}{|c}{$d_2$} & \multicolumn{1}{|c}{$d_3$} & \multicolumn{1}{|c}{$d_4$} \\
\hline
\hline
BBC & 685 & 4 & 5 & 4659 & 4633 & 4665 & 4684 \\
NGs & 500 & 3 & 5 & 2000 & 2000 & 2000 &\multicolumn{1}{|c}{--}\\
WebKB & 203 & 3 & 4 & 1703 & 230 & 230 & \multicolumn{1}{|c}{--}  \\
100leaves & 1600 & 3 & 100 & 64 & 64 & 64 &  \multicolumn{1}{|c}{--} \\
\hline
\end{tabular}
\end{center}
\end{table}

\subsection{Convergence behavior}
\begin{figure}[htbp]
\begin{minipage}[b]{.49\linewidth}
  \centering
\includegraphics[width=1\linewidth]{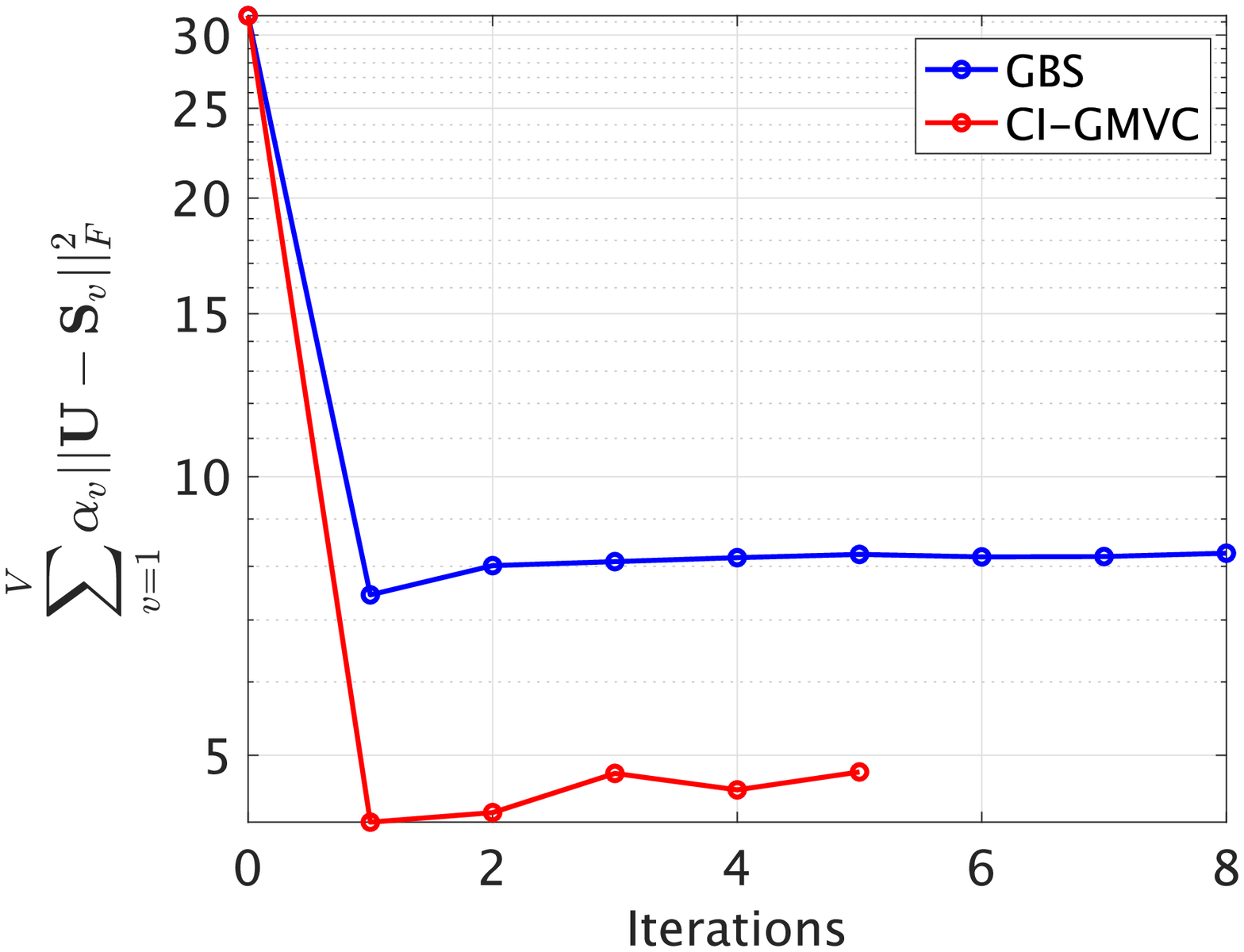}

 {\small (a) WebKB dataset}
\end{minipage}
\begin{minipage}[b]{.49\linewidth}
  \centering
\includegraphics[width=1\linewidth]{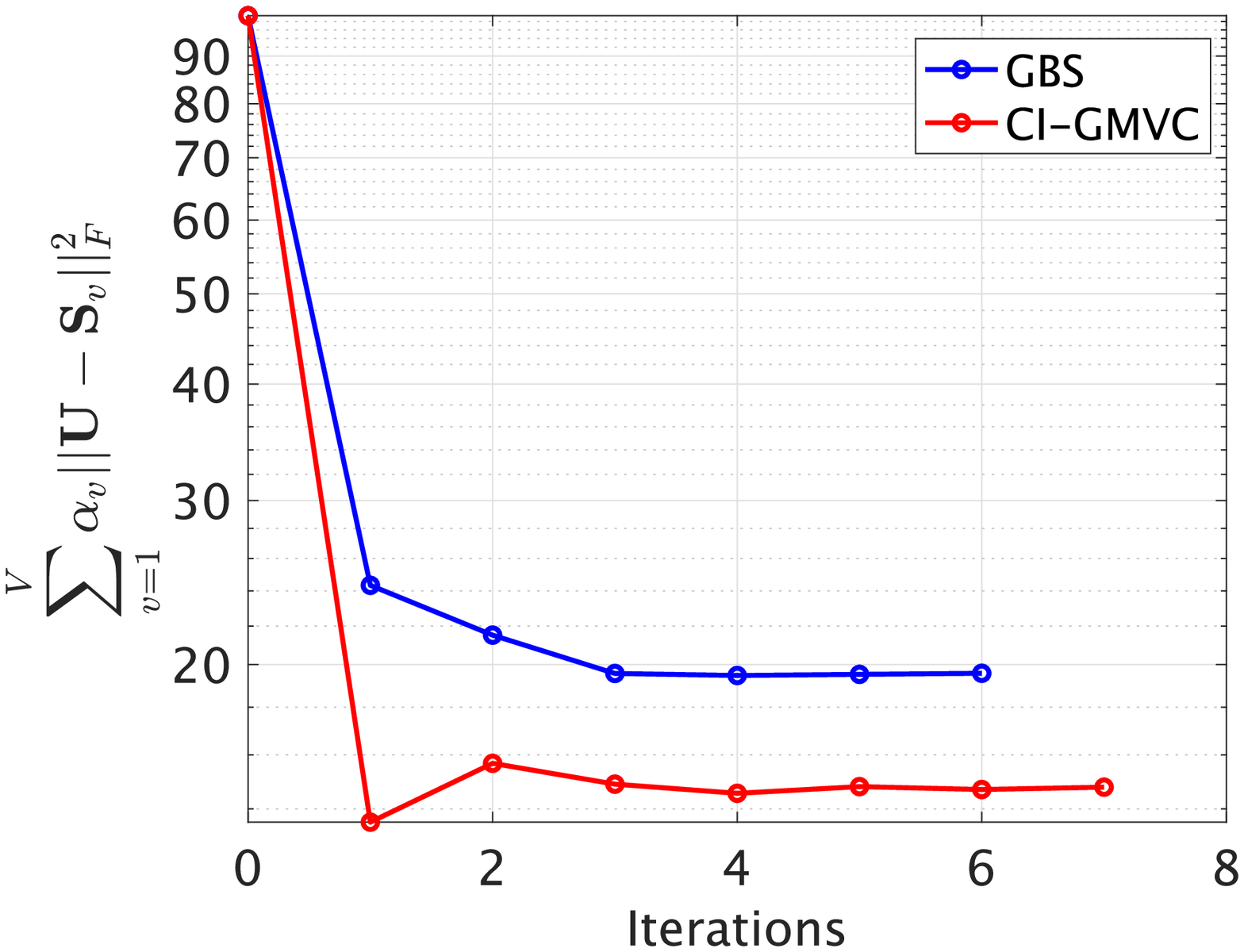}

{\small (b) 100 leaf dataset}
\end{minipage}

\caption{Convergence behaviors of CI-GMVC compared to those of GBS.}
\label{fig:ave_pred_results}
\end{figure}

The objective function in (\ref{Eq:obj_fun}) is not convex on all variables. Therefore, this subsection confirms the convergence behaviors of our proposed CI-GMVC compared with GBS, which outperforms others. For a fair comparison, we evaluate the objective function without regularizers (\ref{Eq:obj_fun}), i.e., $\sum_{v=1}^{V}\alpha_v\|\mat{U}-\mat{S}_v\|_F^2$ in (\ref{Eq:obj_fun_1}), and use the same stopping condition as that used for GBS. As {Fig. \ref{fig:ave_pred_results}} shows, the convergences of CI-GMVC on the WebKB and 100 leaf datasets are faster than those of GBS.

\subsection{Clustering performance}
This subsection presents comparisons of clustering performance. The results of ACC and NMI are shown respectively in {TABLE \ref{tbl:result_ACC}} and {TABLE \ref{tbl:result_NMI}}, where the average scores of the accuracy (ACC) and the normalized mutual information (NMI) are shown. The best performances are presented in bold. Results aside from those of GBS and CI-GMVC are from those in \cite{Wang_KBS_2019_s}. The results demonstrated that the proposed CI-GMVC is comparable to or outperforms other state-of-the-art methods.

\begin{table}[htbp]
\caption{Average clustering performance (ACC).}    
\label{tbl:result_ACC}

\begin{center}
\begin{tabular}{l|c|c|c|c} 
\hline
\multicolumn{1}{c|}{method} & \multicolumn{1}{c|}{\ \ BBC\ \ \ } & \multicolumn{1}{c|}{\ \ NGs\ \ \ \ } &  \multicolumn{1}{c|}{WebKB} &  \multicolumn{1}{c}{100leaves} \\
\hline\hline
CoregSC \cite{Kumar_NIPS_2011} & 47.01& 27.68& 59.70& 77.06\\
\hline
MSC \cite{Xia_AAAI_2014} & 62.32& 31.12& 47.34& 73.79\\
\hline
MGL \cite{Nie_IJCAI_2016} & 53.96& 82.18& 73.84& 69.04\\
\hline
MCGL \cite{Nie_IEEETIP_2018} &35.33 &24.60 & 54.19& 81.06\\
\hline
GBS \cite{Wang_KBS_2019_s} & 69.34& 98.20& 74.38& {\bf 82.44}\\
\hline
CI-GMVC (proposed) & {\bf 70.36}& {\bf 98.40}& {\bf 77.34}&{\bf 82.44} \\
\hline
\end{tabular}
\end{center}

\caption{Average clustering performance (NMI).}   
\label{tbl:result_NMI}

\begin{center}
\begin{tabular}{l|c|c|c|c} 
\hline
\multicolumn{1}{c|}{method} & \multicolumn{1}{c|}{\ \ BBC\ \ \ } & \multicolumn{1}{c|}{\ \ NGs\ \ \ \ } &  \multicolumn{1}{c|}{WebKB} &  \multicolumn{1}{c}{100leaves} \\
\hline\hline
CoregSC \cite{Kumar_NIPS_2011}  & 28.63& \ 8.80& 31.39& 91.65\\
\hline
MSC \cite{Xia_AAAI_2014} &55.31 &\ 9.72 & 22.37& 90.14\\
\hline
MGL \cite{Nie_IJCAI_2016} & 36.97& 83.04& 43.62& 87.53\\
\hline
MCGL \cite{Nie_IEEETIP_2018} & \ 7.41& 10.72& 8.60& 91.30\\
\hline
GBS \cite{Wang_KBS_2019_s}  & 56.27& 93.92& 37.83& 93.43\\
\hline
CI-GMVC (proposed) & {\bf 58.59}& {\bf 94.61}&{\bf 47.01} &{\bf 93.52} \\
\hline
\end{tabular}
\end{center}
\end{table}

\section{Conclusions}
The proposed graph-based multi-view clustering method CI-GMVC particularly incorporates the consistency and the inconsistency structure lying across multiple views. 
Numerical evaluations using several real-world datasets demonstrated the effectiveness of the proposed CI-GMVC. 

\bibliographystyle{unsrt}
\bibliography{supervised_sub_learning,additional}

\end{document}